# Effective Use of Dilated Convolutions for Segmenting Small Object Instances in Remote Sensing Imagery

Ryuhei Hamaguchi    Aito Fujita    Keisuke Nemoto    Tomoyuki Imaizumi    Shuhei Hikosaka
PASCO CORPORATION, Japan
{riyhuc2734, aaitti6875, koetio8807, tiommu4352, saykua3447}@pasco.co.jp

## Abstract

*Thanks to recent advances in CNNs, solid improvements have been made in semantic segmentation of high resolution remote sensing imagery. However, most of the previous works have not fully taken into account the specific difficulties that exist in remote sensing tasks. One of such difficulties is that objects are small and crowded in remote sensing imagery. To tackle with this challenging task we have proposed a novel architecture called local feature extraction (LFE) module attached on top of dilated front-end module. The LFE module is based on our findings that aggressively increasing dilation factors fails to aggregate local features due to sparsity of the kernel, and detrimental to small objects. The proposed LFE module solves this problem by aggregating local features with decreasing dilation factor. We tested our network on three remote sensing datasets and acquired remarkably good results for all datasets especially for small objects.*

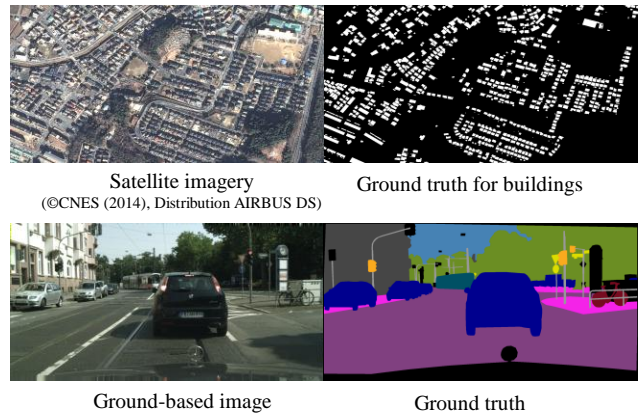

Figure 1. Comparison between satellite imagery from Toyota City Dataset and ground-based image from Cityscapes dataset [35]. The image size is the same in pixel. Corresponding labels are shown on the right.

## 1. Introduction

In recent years, the spatial resolution of satellite imagery has substantially enhanced. Accordingly, the size of objects of interest has become smaller. For example, currently, a spatial resolution of 50 cm is common in commercial satellites such as Pleiades and World View. With such a fine resolution, one can visually distinguish individual instances of small objects (see Figure 1 for building instances). This is important in remote sensing application where number of objects provides valuable information, e.g. the number of newly constructed buildings can be used as economic indicator.

The enhancement of resolution has brought not only the change of target but also a need for methods to utilize rich spatial information. Inspired by the recent deep learning success in computer vision, recent remote-sensing segmentation tasks are addressed by Convolutional Neural Networks (CNNs). Although solid improvements have been made by CNNs [8, 10, 12, 14, 15], most of the previous works directly employ modern CNN architectures with minor modifications and they do not consider a specific difficulty of remote sensing imagery.

Figure 1 intuitively explains the difficulty. In this figure, representative samples are compared between Pleiades satellite imagery and ground-based image (from Cityscapes dataset [35]). From the two images we can observe the following differences. (i) Size of objects: compared to the ground-based image, the objects in the satellite imagery are significantly smaller[1]. (ii) Layout of objects: in satellite imagery, the objects are densely located. In light of these differences, designing a dedicated architecture is obviously needed for remote-sensing segmentation rather than directly employing modern CNN architectures.

To segment such a tight crowd of small objects, one of the most important elements is context in an image. [26] showed the importance of context for CNNs to recognize small objects. In CNNs, large context is acquired by subsampling layers. Although subsampling layers are helpful to expand the receptive field, they ignore the other important element: resolution. Resolution is important to resolve a tight crowd of small objects. Nonetheless, by subsampling layers resolution of features is gradually lost through layers of a network. The resulting coarse features can miss the details of small objects that are difficult to recover even with efforts such as skip connections [1, 5] or hypercolumns [6, 21]. Thus, we need a specific method to expand the receptive field without losing resolutions.

---

[1] For example, the mean bbox size of annotated objects is 26 pixels for our satellite imagery dataset (Section 4.3) vs. 168 pixels for PASCAL dataset.

As the promising method, [3] proposed dilated convolutions. In dilated convolutions, the alignment of the kernel weights is expanded by dilation factor. By increasing this factor, the weights are placed far away at given intervals (i.e., more sparse), and the kernel size accordingly increases. Therefore, by monotonously increasing the dilation factors through layers, the receptive field can be effectively expanded without loss of resolution. Actually dilated convolutions work quite well in current computer-vision papers [2, 3, 7]

Nevertheless, we highlight that a naive application of dilated convolutions does not always improve performance. Specifically, aggressively increasing dilation factors fails to aggregate local features of small objects. This is a side-effect of increased interval of the kernel weights, i.e. the increased sparsity of the kernel (explained in Section 3). This means that whereas increasing dilation factors is important in terms of resolution and context, it can be detrimental to small objects. This is especially undesirable for remote sensing scenario. While CNNs equipped with increasingly dilated convolutions are all the rage in modern vision studies, segmentation of small objects should be addressed otherwise.

We solve this problem by simply going against the tide—decreasingly dilated convolutions. To this end, we propose a novel module, which we call Local Feature Extraction (LFE) module (Figure 2). The LFE module consists of several convolutional layers with decreasing dilation factors. Specifically, we attach the LFE module on top of increasingly dilated convolutions. Such a combination is preferable: local features are aggregated as the kernel weights get denser through the LFE module. In other words the LFE module acts as rescue of increasingly dilated convolutions.

We comprehensively evaluate our method on three remote sensing datasets. Across all the datasets, the proposed model outperforms state-of-the-arts such as U-Net [5] and Deeplab [2], especially for small objects. To analyze the effect of the LFE module we conduct the effective receptive field (ERF) analysis [27] and find that the LFE module smoothen grid-like ERF pattern that appears in the trained models with dilated convolution.

## 2. Related work

Semantic segmentation is a task to assign each pixel in an input image a semantic category. Since FCNs [1] have extended well studied classification network to dense pixel labeling settings, large advances have been made in this field. One challenging problem concerning semantic segmentation is how to precisely localize objects. Simple extension of classification networks fails to extract clear boundaries because of spatially abstracted coarse features. [1] approached this problem by integrating multi resolution prediction maps from different stages of their network. Other approach is based on encoder-decoder architecture. In [4], low resolution semantic features are first extracted in encoder part, then spatial resolution of the features are recovered in decoder part using the selected position of max pooling layer in the encoder as cue. Instead of using max pooling position in decoding process, [5] progressively refine features skipping and combining low level features in their encoder network. This notion of integrating multi resolution features is also common in [6, 21]. Another approach is based on dilated convolution. In [3] dilated convolutions were utilized to effectively expand receptive field without losing resolution. The contemporaneous works [32, 33] also observed the same problem of dilated convolutions as we pointed out (especially the first part of the problem which we will explain in Section 3.3). To remedy the problem, they proposed the successive use of decreasing dilation factors. This approach is conceptually the same as ours, but the objective is different: they aimed to improve "semantic segmentation performance for ground-based image" whereas we aim to improve "instance-level segmentation performance for small objects in remote sensing imagery".

In remote sensing domain, semantic segmentation of satellite or aerial imagery is also well studied. Most works follow the architectural improvements on computer vision community such as the works utilizing FCN [10, 13], skip connections [11], encoder-decoder architectures [12] or dilated convolutions [14]. Among these, [14] is close to our work as they also used dilated convolutions to avoid down sampling. However, they also used max-pooling layers with stride of 1 just after each dilated convolutions, which decreases actual resolution of the extracted feature maps. In contrast, our method uses no pooling operation and keeps the same resolution as the inputs. In [13] segmentation accuracy for small objects was improved by using class balanced loss function. In [16], pixels in input images were classified according to the distance from boundary of each object instance which results in precise localization of object boundary. The focus of these two works is design of loss functions and our method is orthogonal to these works.

The goal of our work is not only the semantic segmentation of remote sensing imagery, but also the detection of individual object instances. Such task is classified as instance-aware semantic segmentation. In previous works, the task is mainly approached by two step pipelines: the object mask proposal step and the subsequent classification step. Since this task was first proposed in [17], several improvements have been made by utilizing shared CNN features for individual proposals [18], training multi-task CNN for object proposal and classification [19] and making object proposals in FCN manner [20]. Some of these methods rely on object proposal methods. Especially in [23], CNN based object proposal method was proposed. The common focus of these works is how to resolve occlusions often encountered in ground-based images. However, compared to ground-based images, occlusion is not serious problem in remote sensing imagery. Instead, focus of our work is how to precisely segment small object instances often encountered in remote sensing imagery.

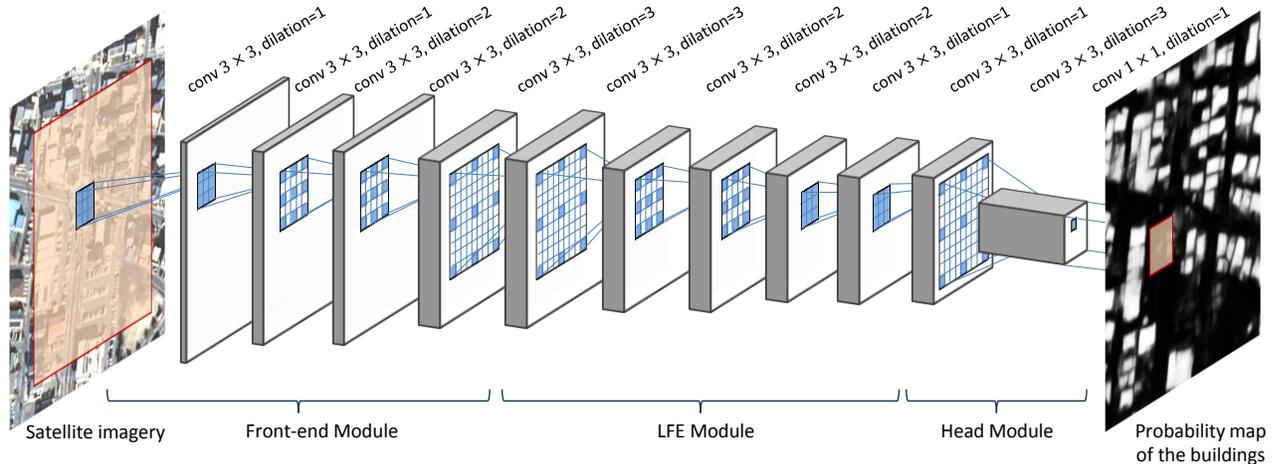

Figure 2. Overview of the proposed network architecture. In each layer, filter kernels are depicted in square with grid pattern. The blue cells in dilated kernel represent valid weights and blank cells represent invalid region. As shown, in case of kernels with dilation factor of 2, valid weights align by interval of 1 and in case of 3, they align by interval of 3.

## 3. Proposed Method

### 3.1. Overview of the Proposed Method

As pointed out in [26], context information matters to detect small objects. Even humans cannot recognize a small building in a satellite imagery patch without context information such as roads, cars or other buildings. Also, a higher spatial resolution is crucial. In coarse resolution, small objects can be over-segmented into a single mask, or missed. Thus, we should pay attention to both context and resolution.

Figure 2 shows a schematic of the proposed segmentation model: front-end module, local feature extraction (LFE) module and head module. All modules are designed to keep resolution by using dilated convolution layers. The role of each module is different. The front-end module is designed to extract features that cover large context, and thus the dilation factors are gradually increased (Section 3.2). Conversely, the subsequent LFE module is dedicated to aggregating local features scattered by the front-end module.

Thus, the LFE module has the specific structure of decreasing dilation factors (Section 3.3). Finally, the head module outputs a probability map with the same resolution as input. The module is the convolution version of fully connected layers of classification networks such as VGG.

As post-processing, the output probability maps are used to acquire mask proposals. This is simply done by thresholding (Section 3.4).

### 3.2. Front-end module

The role of front-end module is to aggregate large context. In many CNN models, subsampling layers are effectively used to enlarge the receptive field size. However, subsampling layers decrease the spatial resolution of learned features. One simple approach of eliminating subsampling layers fails because the number of parameters explodes for maintaining the same receptive field size as before.

In order to satisfy both of a large receptive field and a high spatial resolution, we adopt dilated convolutions [3]. The dilated convolutions enlarge receptive field while maintaining resolution. As shown in Figure 2, dilated convolutions utilize specific kernels with sparsely aligned weights. Both of the kernel size and the interval of sparse weights expand exponentially with dilation factor. By increasing dilation factor, receptive field is also expanded exponentially by large kernel. Although previous works commonly use dilated convolution in a few layers near the output, we take more drastic approach. Specifically, we eliminate all subsampling layers of front-end module and use dilated convolution instead. Though this is effective for small buildings, there are two problems concerning sparsity in dilated kernels. The problems are explained in the next section.

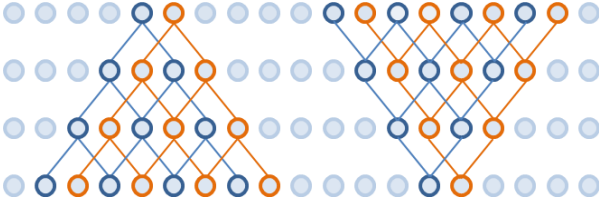

Figure 3. Description of non-overlap field of view of adjacent two units on top-most layer (**Left**) and non-overlap area of influence of adjacent two units on bottom-most layer (**Right**).

### 3.3. Local feature extraction module

The role of LFE module is to solve problems of front-end module. Specifically, aggressive application of dilated convolution causes two problems, (1) spatial consistency between neighboring units becomes weak and (2) local structure cannot be extracted in higher layer. In this subsection, we first describe these two problems in detail. Then explain how the LFE module solves the problems.

**Problem on spatial inconsistency:** Suppose 1D convolutions with the kernel size of 2 and the dilation factor of 2. In left side network in Figure 3, the blue unit on top most layer is affected by blue units in lower layers, and so as orange one. These same color units compose information pyramids which defines the field of view of top most units. We can see that information pyramids of two adjacent units do not overlap due to the sparse connections of the dilated kernels. In the case of the dilation factor of 2, two neighboring units have non-overlap information pyramids, and as we increase the dilation factor, number of neighboring units which have non-overlap information pyramids grows larger. In the case of front-end module with increasing dilation factors, information pyramids gradually branch off from input to output. As long as dilation factor increases, they never overlap again in higher layers. In this way, field of views of neighboring units in output layer are only slightly overlap at lower layers. As confirmed in experiment section, this causes spatial inconsistency between neighboring units and causes serious jaggy patterns in final output maps.

**Problem on local structure extraction:** In Figure 3, right side network also illustrates information pyramid but upside down from left side. In this case, information pyramid defines area of influence from bottom most unit. Again, information pyramids do not overlap for two adjacent units in bottom most layer. All units in top most layer receive information from either of the two units, but not both. This means that all units in top most layer are unaware of local structure inside the two units. As well as the first problem, non-overlap region grows larger as dilation factor increases.

|  | Front-S | Front-S+D | Front-S+D+LFE |
|---|---|---|---|
| Front | conv-n64-k3-d1<br>conv-n64-k3-d1<br>max pooling<br>conv-n128-k3-d1<br>conv-n128-k3-d1<br>max pooling<br>conv-n256-k3-d1<br>conv-n256-k3-d1<br>conv-n256-k3-d1 | conv-n64-k3-d1<br>conv-n64-k3-d1<br><br>conv-n128-k3-d2<br>conv-n128-k3-d2<br><br>conv-n256-k3-d3<br>conv-n256-k3-d3<br>conv-n256-k3-d3 | conv-n64-k3-d1<br>conv-n64-k3-d1<br><br>conv-n128-k3-d2<br>conv-n128-k3-d2<br><br>conv-n256-k3-d3<br>conv-n256-k3-d3<br>conv-n256-k3-d3 |
| LFE |  |  | conv-n256-k3-d3<br>conv-n256-k3-d3<br>conv-n256-k3-d3<br>conv-n256-k3-d2<br>conv-n256-k3-d2<br>conv-n256-k3-d1<br>conv-n256-k3-d1 |
| Head | conv-n1024-k9-d1<br>conv-n1024-k1-d1<br>deconv-n2-k16-d1 | conv-n1024-k7-d3<br>conv-n1024-k1-d1<br>conv-n2-k1-d1 | conv-n1024-k7-d3<br>conv-n1024-k1-d1<br>conv-n2-k1-d1 |

Table 1. Detailed architectures of the networks. In the table, "conv-n(a)-k(b)-d(c)" represents a convolutional layer with b×b kernel, dilation factor of c and output number of feature maps of a. "maxpooling" represents max-pooling layer with kernel size of 2×2 and stride of 2.

In case of front-end module with increasing dilation facto r. If target objects are large enough to recognize their local structure from features inside the object, this is not the problem. In this case, local structure can be completely extracted by denser kernels at lower layers. However, in case of small objects, some of local structures need to be extracted at higher layers because large context is needed to recognize them. However, with increasing dilation factor, higher layers cannot extract local structure because of non-overlap information pyramid.

**Local feature extraction module:** To handle these two problems, we propose local feature extraction module (LFE) with *decreasing* dilation factor. The idea is that the main cause of the problems is increasing dilation factor. If we attach structure with decreasing dilation factor after increasing one, information pyramids of neighboring units can be connected again. Thus, decreasing structure gradually recovers consistency between neighboring units and extracts local structure in higher layer. In experiments section, the LFE module is shown to be effective especially for small objects.

### 3.4. Post-processing

In our model, mask proposals for individual object instances are acquired by simply thresholding the output probability map. Then, for each mask, an object score is computed as a mean of probability values inside the mask.

| | pixel F1 | APr | APvol | AR | AR Very small | AR Small | AR Mid | AR Large | AR Very large |
|---|---|---|---|---|---|---|---|---|---|
| U-Net [5] | 62.7% | 28.4% | 24.9% | 21.6% | 0.8% | 21.9% | 36.8% | 34.5% | 27.5% |
| FCN-8s [1] | 56.2% | 5.8% | 9.8% | 4.2% | 0.0% | 2.1% | 8.4% | 21.5% | 17.5% |
| Deeplab [2] | 61.7% | 11.3% | 15.4% | 7.3% | 0.0% | 4.7% | 14.4% | 24.8% | 24.4% |
| Sherrah [14] | 62.0% | 22.7% | 21.3% | 17.0% | 0.2% | 15.1% | 30.8% | 34.0% | 32.5% |
| Front-S | 58.6% | 26.5% | 23.1% | 18.9% | 0.7% | 18.0% | 33.2% | 29.9% | 33.8% |
| Front-S+D | 60.1% | 30.3% | 25.3% | 21.9% | 0.9% | 23.5% | 36.2% | 29.2% | 26.5% |
| Front-S+D+Large | 59.0% | 33.1% | 27.0% | 25.4% | 1.2% | 27.5% | 41.8% | 33.7% | 40.0% |
| Front-S+D+Large+CRF | 62.5% | 32.6% | 26.4% | 23.9% | 0.9% | 25.9% | 39.3% | 31.3% | **42.5%** |
| Front-S+D+LFE | 60.1% | **33.5%** | **27.7%** | **25.8%** | **1.5%** | **28.1%** | **42.0%** | 34.5% | 33.1% |
| Front-S+D+LFE+CRF | **63.6%** | 32.4% | 26.9% | 24.3% | 1.1% | 26.3% | 40.0% | 32.6% | 28.1% |
| Front-L | 59.7% | 26.3% | 22.9% | 18.5% | 0.6% | 16.5% | 33.6% | 33.4% | 37.3% |
| Front-L+D | 59.1% | 28.4% | 23.6% | 22.8% | 1.1% | 23.4% | 38.7% | 32.4% | 28.1% |
| Front-L+D+Large | 60.4% | 27.7% | 24.0% | 21.8% | 1.1% | 21.0% | 37.9% | **36.1%** | 36.9% |
| Front-L+D+LFE | 61.1% | 31.4% | 26.4% | 24.0% | 1.3% | 24.4% | 40.9% | 30.1% | 24.4% |

Table 2. The accuracy of the models on Toyota City Dataset. AP$^r$ is computed at IoU over 0.5. We use our own implementation for U-Net model and the model proposed by Sherrah et al.

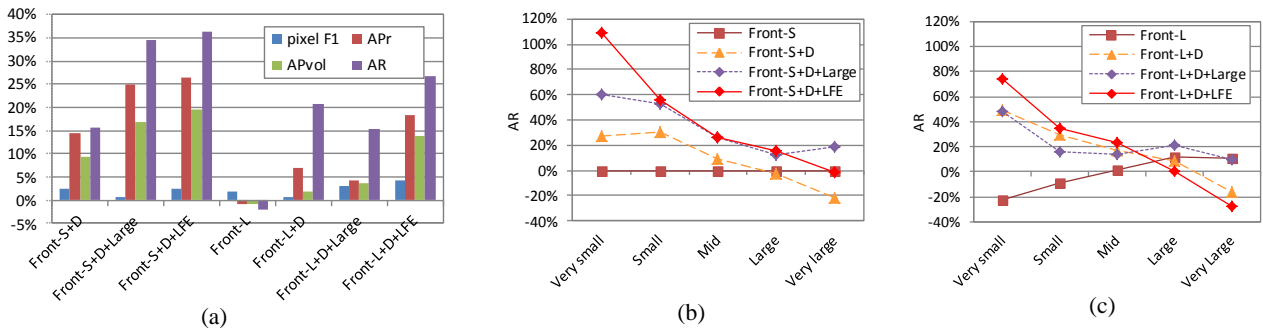

Figure 4. Relative improvements compared to baseline model (Front-S) on Toyota City Dataset. Results on each metric for whole test set are depicted in (a), and results on AR metric for each object sizes are depicted in (b) and (c).

Though very simple, this method works well for remote sensing imagery where, differently from ground-based images, occlusion between objects is not so serious. Even if serious, the occlusion problem could be addressed by integrating our modules into previously proposed instance-aware semantic segmentation pipelines (e.g., [20, 22, 24]).

## 4. Experiments

In this section, we evaluate our method on three datasets. The first one, Toyota City Dataset, is used to establish and validate our method. The other two, Massachusetts Buildings Dataset [8] and Vaihingen Dataset [31], are used to benchmark our method with previously proposed segmentation methods.

### 4.1. Evaluation metric

In our experiments, we use AP$^r$, AP$^r_{vol}$ [17] and AR [25] to evaluate our method. These metrics are commonly used evaluation metrics for instance-aware semantic segmentation and mask proposal generation task. In AR evaluation, it is common to use fixed number of proposals. However, fixed number of proposals is not suitable in this case since in remote sensing imagery, the number of objects changes drastically across scenes (e.g. buildings in urban scene and rural scene) Instead, we compute AR for every proposals and use AP$^r$ and AP$^r_{vol}$ to consider false alarms.

### 4.2. Basic setup for experiments

Our experiments comprise three axes: the size of field of view (small FOV or large FOV), expansion strategy of receptive field (pooling or dilation) and use of LFE module (with or without). First, in terms of FOV, two types of pooling-based front-end models are trained with different field of view. The architecture of both models is based on VGG-16 [28], but higher layers are eliminated. Specifically, only layers bellow third pooling layer of VGG-16 are used for small FOV (Front-S) and layers bellow fourth pooling layer are used for large FOV (Front-L). We set Front-S as baseline of our experiments. Second, to validate the effect of dilated convolution, the dilated versions of each front-end model are trained (Front-S+D, Front-L+D).

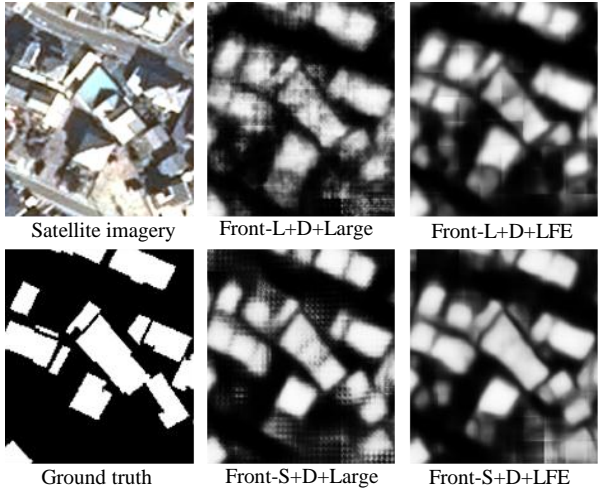

Figure 5. Example output probability maps for different models. We can see jaggy patterns for the models without LFE module (**Mid column**), but they disappear for the models with LFE module (**Right column**). (Satellite imagery: ©CNES (2014), Distribution AIRBUS DS)

Specifically, we eliminate all pooling layers from pooling-based front-end models and convert convolution to dilated convolution to keep the same FOV. Thirdly, to validate the effect of the proposed LFE module, we attach the LFE module to the dilated front-end and train end-to-end (Front-S+D+LFE, Front-L+D+LFE). In this case, to ensure fairness in terms of parameter size, we train counterparts (Front-S+D+Large, Front-L+D+Large) that have the same number of parameters as the models with LFE module. The only, but important difference is that the counterpart models do not have decreasing dilation factor. Instead, the dilation factors of the corresponding layers are kept equal. Detailed architectures of main models are shown in Table 1. In all of the models, convolutional layers except the last one are followed by ReLU activations. The last convolutional layer is followed by softmax layer to output probability map. All networks take 76×76 patches as input and output probability maps for center area of size 16×16.

### 4.3. Experiments on Toyota City Dataset

**Dataset:** The Toyota City Dataset is composed of satellite imagery around Toyota City, Japan. The images were acquired by Pleiades satellite in 2014. Training and test data covers roughly 200 km$^2$ and 20 km$^2$, each containing 100,000 and 15,000 buildings. Image resolution is 50 cm and RGB bands are used. Labels are provided for two classes: building or non-building for each pixel. For training, patches are randomly cropped and augmented by random rotation. Then, we balance the samples by the number of building pixels in the sample. A total of 400k patches are collected in this way for training.

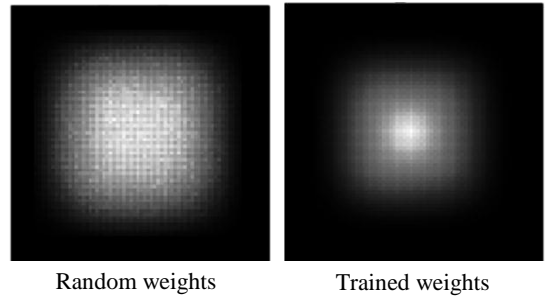

Figure 7. ERF visualization for dilated front model (Front-S+D). **Left**: weights of the network are randomly initialized using [29]. **Right**: weights are trained with Toyota City Dataset.

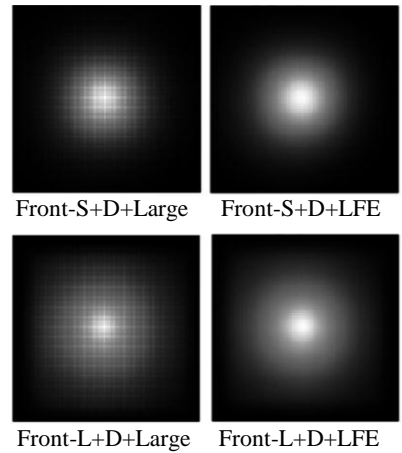

Figure 6. ERF visualization results for the trained models. The grid like ERFs in dilated front-end model (**Left**) are successfully smoothed by the LFE module (**Right**).

**Results of experiment:** As we see in Table 2 and Figure 4 (a), LFE module performs the best in instance metric, and dilated front-end modules perform better than pooling. In terms of FOV of front-end module, small FOV performs better than large FOV. To analyze sensitivity with respect to object size, we divided buildings into five classes by its extent: Very small (0-100 pixels), Small (100-400 pixels), Mid (400-1,600 pixels), Large (1,600-6,400 pixels) and Very large (over 6,400 pixels). Then we evaluate AR for each size of the buildings. Figure 4 (b) and (c) shows relative AR improvement from Front-S model. As we can see, dilated front-end module and LFE module shows remarkable performance gain for small buildings. The example output probability map for a large test scene is shown in Figure 10.

We also applied dense-CRF [34] for Front-S+D+Large and Front-S+D+LFE. While pixel F1 scores are almost equally improved for the both models, instance level performances are degraded. One reason may be that CRFs are not suitable to small objects because they usually have weak contrast and ambiguous boundaries, which is difficult for CRFs to separate individual instances.

|  | pixel F1 | APr | APvol | AR | AR Very small | AR Small | AR Mid | AR Large | AR Very large |
|---|---|---|---|---|---|---|---|---|---|
| U-Net [5] | 94.1% | 61.9% | 49.3% | 32.9% | 27.0% | 36.9% | 38.2% | **45.0%** | 52.9% |
| FCN-8s [1] | 93.1% | 26.1% | 29.5% | 12.3% | 10.6% | 16.3% | 27.4% | 39.2% | 45.4% |
| Deeplab [2] | 89.7% | 7.2% | 13.0% | 4.4% | 1.5% | 4.2% | 18.7% | 28.5% | 34.3% |
| Sherrah [14] | 93.0% | 51.9% | 43.0% | 26.5% | 21.9% | 29.3% | 33.6% | 37.8% | 38.2% |
| Mnih-CNN [8] | 91.5% | — | — | — | — | — | — | — | — |
| Saito-CNN-MA [9] | **94.3%** | — | — | — | — | — | — | — | — |
| Front-S | 93.3% | 53.7% | 43.8% | 27.1% | 22.2% | 30.2% | 33.3% | 39.0% | 41.7% |
| Front-S+D+Large | **94.3%** | 62.7% | 49.5% | 33.8% | 28.2% | 37.9% | **38.5%** | 42.9% | 38.7% |
| Front-S+D+LFE | 93.4% | **64.0%** | **50.3%** | **35.0%** | **28.9%** | **39.7%** | 38.2% | 43.1% | 42.1% |

Table 4. The accuracy of the models on Massachusetts Buildings Dataset. We use our own implementation for U-Net model and the model proposed by Sherrah et al.

|  | pixel F1 | APr | APvol | AR |
|---|---|---|---|---|
| U-Net [5] | 76.7% | 72.9% | 61.5% | 52.0% |
| FCN-8s [1] | **81.0%** | 52.8% | 47.2% | 32.5% |
| Deeplab [2] | 79.3% | 66.2% | 54.8% | 43.6% |
| Sherrah [14] | 77.7% | 57.8% | 48.0% | 33.4% |
| Front-S | 80.0% | 66.3% | 54.6% | 46.9% |
| Front-S+D+Large | 77.8% | 77.5% | 65.1% | **57.5%** |
| Front-S+D+LFE | 77.9% | **77.6%** | **65.7%** | 56.5% |

Table 3. The accuracy of the models on Vaihingen Dataset. We use our own implementation for U-Net model and the model proposed by Sherrah et al.

**Effect of LFE module:** Though simply replacing pooling layer with dilated convolution improves performance on $AP^r_{vol}$, the performance gain is relatively low for the front-end with large FOV (i.e. +2.2% in case of Front-S and Front-S+D, but +0.7% in case of Front-L and Front-L+D). This result can be explained by the spatial inconsistency problem explained in section 3. Since Front-L+D have larger dilation factor than Front-S+D, the spatial inconsistency between adjacent features become stronger and this might have hurt performance. Note that the performance gain by using LFE module is remarkably higher for large FOV (i.e. +0.7% in case of Front-S+D+Large and Front-S+D+LFE, and +2.4% in case of Front-L+D+Large and Front-L+D+LFE). This implies that the proposed LFE module successfully solves the spatial inconsistency problem. We can also see the effect of LFE module in output probability maps in Figure 7. In the figure, we can see harmful jaggy patterns caused by the spatial inconsistency problem in the output of dilated front-end models. Note that these patterns are smoothed in models with LFE module.

**Analysis in terms of metric:** In Figure 4 (a), we can see little relation between pixel level metric and instance level metrics. For example, while pixel level F1 measure of Front-S+D+Large only slightly better than Front-S (+0.4%), a large improvement is achieved in instance level metrics (e.g. +7.0% in $AP^r$). These results reflect the fact that small objects have little impact on pixel level metric. In contrast, we can see positive relation between $AP^r_{vol}$ and AR. There

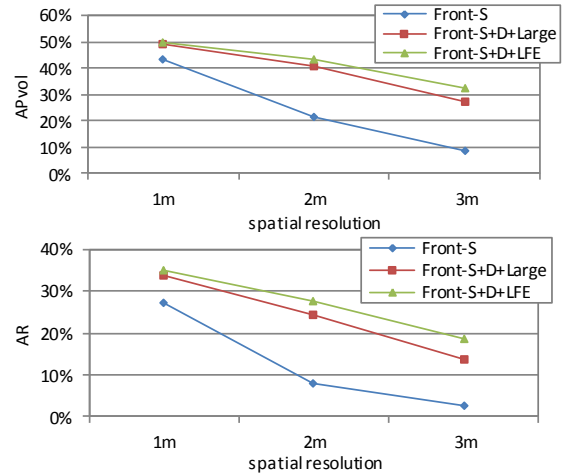

Figure 8. Results for sensitivity analysis on input resolution on Massachusetts Buildings Dataset. Results are shown for $AP^r_{vol}$ (**Upper**) and for AR (**Lower**)

seems to be a tendency that models with high AR also achieves low false alarm rate.

**ERF Analysis:** To further analyze the effect of LFE module, we visualize effective receptive field (ERF) of the models. As is done in [27], visualization process is as follows. First set gradient 1 for center unit in output map and 0 for others. Then this gradient map is back-propagated to compute input gradients. We compute input gradients over all patches in our validation set and average their absolute value to form ERF map. In Figure 7, we compare the ERF between random weight case and trained weight case for our dilated front-end module (Front-S+D). In former case, weights are initialized using [29] and in latter case, they are trained with Toyota City Dataset. In random weight case, rectangular shape can be seen in ERF which is consistent to the result of [27]. However, to our surprise, a systematic grid pattern appears in trained weight case. This grid like ERF is problematic since local structure smaller than the grid scale cannot be captured in the output. One explanation of this grid pattern is as follows: sparse connections of dilated kernels at higher layer propagate spatially sparse gradient signals. In random weight case, these gradients are smoothed by uniformly distributed weights in dense kernels at lower layer. However, in trained weight case, kernels in lower layers may have relatively centered distribution and they cannot smooth sparse gradients enough which results in grid like ERF. This means that, in our trained dilated front-end module, local information are not sufficiently captured in lower layers.

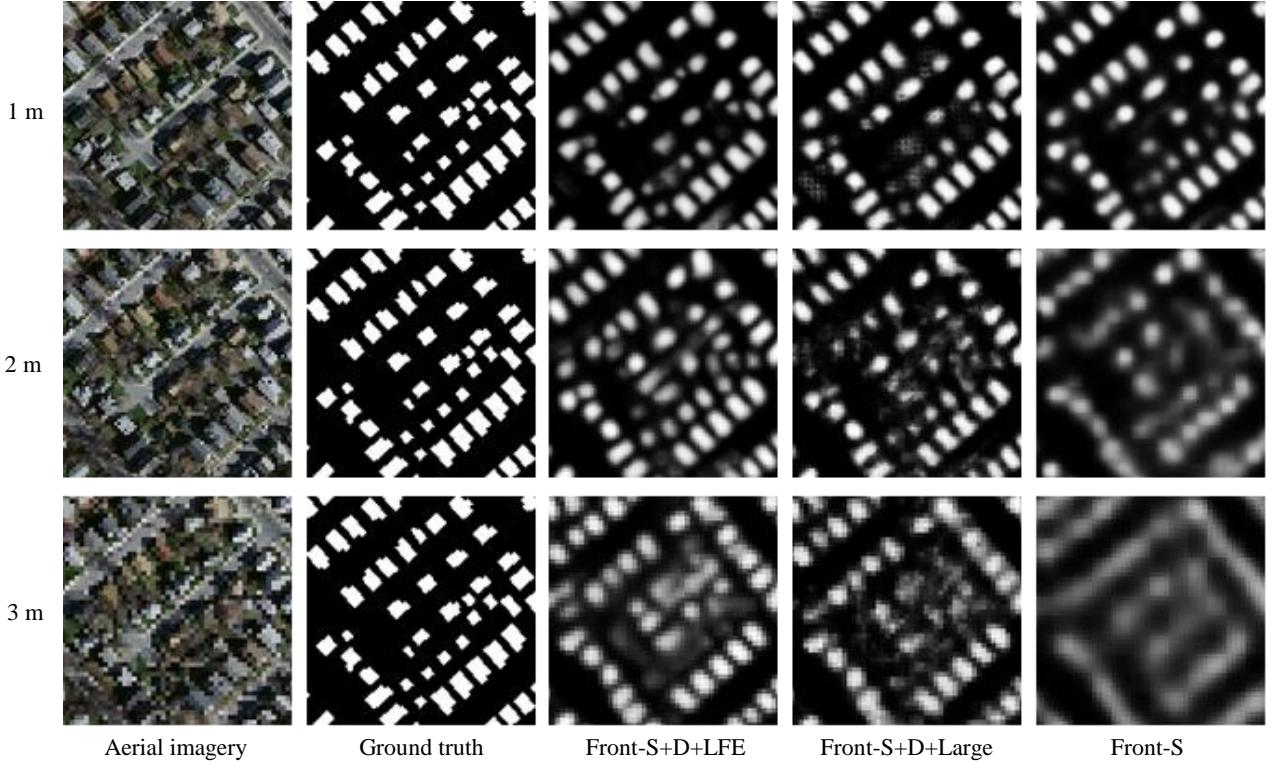

Figure 9. Example of output probability maps for different resolution of Massachusetts Buildings Dataset. From top to bottom, the resolution of the datasets is down sampled. As we see, the model with LFE module can detect small objects even with resolution of 3 m (**Third column**). In contrast, the pooling-based front-end model over estimates many buildings and fails to detect individual buildings at lower resolution (**Fifth column**).

To see the effect of the LFE module, we conducted the same ERF analysis for trained models with and without LFE module. As shown in Figure 6, grid patterns in dilated front-end modules are smoothed in the models with LFE module. This means that the proposed LFE module successfully grasp local information missed in lower layers.

### 4.4. Experiments on Massachusetts Buildings Dataset and Vaihingen Dataset

**Experimental Setup:** Massachusetts Buildings Dataset [8] is composed of 1 m spatial resolution aerial imagery with RGB bands. The dataset covers roughly 340 $km^2$ with 194,070 buildings for training and 23 $km^2$ with 15,261 buildings for testing. Following [9], we randomly crop training patches. We augment each patch by random rotation and acquired 400k training patches in total. The other dataset, Vaihingen Dataset, is provided by Commission III of the ISPRS [31]. The dataset is composed of 9 cm spatial resolution aerial imagery. We use near infrared, red and green bands and do not use digital surface model (DSM). Dataset includes 16 labeled scenes which cover roughly 0.6 $km^2$. Following previous works [11, 14, 15], we use 5 scenes (IDs: 11, 15, 28, 30, 34) for validation and remaining 11 scenes for training. Labels are provided for 6 classes: impervious surface, building, low vegetation, tree, car and clutter/background. In this experiment, we set car class as our target and use only car labels for training and testing. For both of datasets we modify the architecture of the LFE module to have four convolutional layers (First two have kernel size of 3 and dilation factor of 2 followed by the other two with kernel size of 3 and dilation factor of 1). This modification is just a tuning and do not affect tendency of the result.

**Comparison to other methods:** In Table 4 and Table 3, we compared the performance of our models with previously proposed models. In most of instance level metrics, the proposed model with LFE module performs the best for both datasets. In pixel level metric, Front-S performs competitive to previous works which ensures our baseline.

We set boundary margin of 3 pixels to evaluate pixel F1 following previous works [8, 9].

**Sensitivity analysis on input resolution:** In order to analyze the sensitivity of our methods on input resolution, we establish two datasets with different resolution: 2 m and 3 m by down sampling all images in original Massachusetts Buildings Dataset. These datasets are more challenging because size of the buildings becomes significantly smaller and local structure such as edge of buildings becomes more abstract (see aerial images in Figure 9). For these datasets, the baseline model and its dilated version (with and without LFE) are trained and tested. Figure 8 shows how performances of the models change according to the resolution. As we see, the model with LFE module performs the best for all resolution datasets. More importantly, the performance improvement by the LFE module is more remarkable at lower resolution, which shows the effectiveness of the LFE module to small objects. In contrast, the performances of pooling-based front-end module (Front-S) degrade rapidly, showing the importance of feature resolution for small objects. Examples of output probability maps are shown in Figure 9. Again in this figure, we can see effectiveness of the model with LFE module. This is worth noticing that this analysis is also important in application aspect because several meter resolution imagery have becoming next target product in the field of earth observation satellite. The methods to recognize ground objects from such resolution will have significant importance in the near future.

## 5. Conclusion

In this paper, we have presented novel network architecture based on dilated convolution to precisely segment crowded small object instances in remote sensing imagery. In particular, we have pointed out the problem in conventional use of dilated convolution, and proposed architecture to solve the problem.

Our method shows remarkable effectiveness for small object instances in three remote sensing dataset, suggesting promising application to various remote sensing tasks.

Finally, our idea is not limited in remote sensing tasks and expected to be effective where crowded small instances matters: segmentation of cells in biomedical domain, crowd counting, pedestrian detection and more.

## Acknowledgements

The Vaihingen data set was provided by the German Society for Photogrammetry, Remote Sensing and Geoinformation (DGPF) [31]:
http://www.ifp.uni-stuttgart.de/dgpf/DKEP-Allg.html.

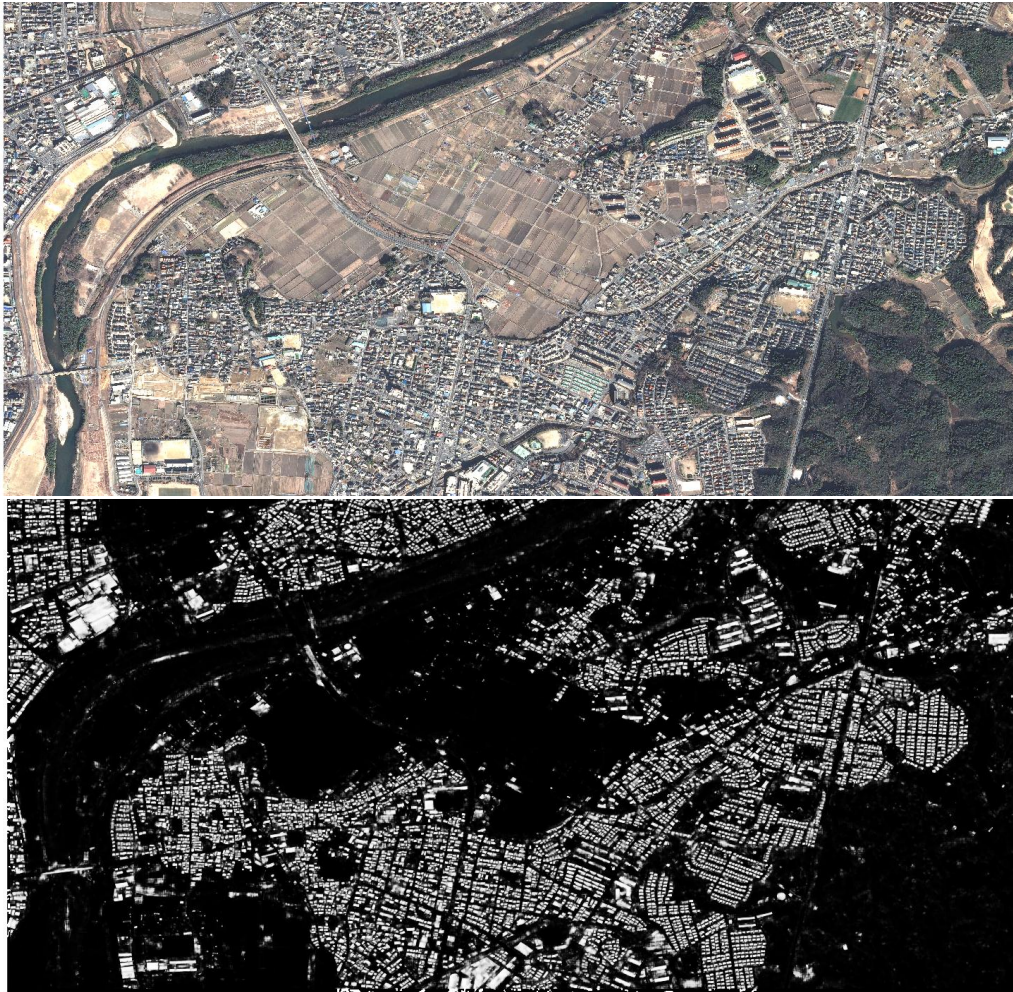

Figure 10. The result of building extraction for large test scene in Toyota City Dataset. **Upper**: Satellite imagery (©CNES (2014), Distribution AIRBUS DS). **Lower**: The corresponding output probability map of proposed model (Front-S+D+LFE)